\documentclass{ecai} 

\usepackage{multirow}
\usepackage{latexsym}
\usepackage{amssymb}
\usepackage{amsmath}
\usepackage{amsthm}
\usepackage{dsfont}
\usepackage{booktabs}
\usepackage{enumitem}
\usepackage{graphicx}
\usepackage{color}
\usepackage{array}
\usepackage{makecell}

\usepackage{bm}

\newcolumntype{P}[1]{>{\centering\arraybackslash}p{#1}}

\newcommand{\BibTeX}{B\kern-.05em{\sc i\kern-.025em b}\kern-.08em\TeX}

\begin{document}


\begin{frontmatter}




\title{VANER: Leveraging Large Language Model for Versatile and Adaptive Biomedical Named Entity Recognition}


\author[A,F]{\fnms{Junyi}~\snm{Bian}\footnote{Equal contribution.}}
\author[B]{\fnms{Weiqi}~\snm{Zhai}\footnotemark}
\author[G]{\fnms{Xiaodi}~\snm{Huang}}
\author[B]{\fnms{Jiaxuan}~\snm{Zheng}}
\author[B,C,D,E,F]{\fnms{Shanfeng}~\snm{Zhu}\thanks{Corresponding Author. Email: zhusf@fudan.edu.cn}} 

\address[A]{School of Computer Science, Fudan University, Shanghai 200433, China}
\address[B]{Institute of Science and Technology for Brain-Inspired Intelligence, Fudan University, China}
\address[C]{Key Laboratory of Computational Neuroscience and Brain-Inspired Intelligence \\ (Fudan University), Ministry of Education, Shanghai 200433, China}
\address[D]{MOE Frontiers Center for Brain Science, Fudan University, Shanghai 200433, China}
\address[E]{Zhangjiang Fudan International Innovation Center, Shanghai 200433, China}
\address[F]{Shanghai Key Lab of Intelligent Information Processing, Fudan University, Shanghai 200433, China}
\address[G]{School of Computing and Mathematics, Charles Sturt University \\ Albury, NSW 2640, Australia}




\begin{abstract}




Prevalent solution for BioNER involves using representation learning techniques coupled with sequence labeling. However, such methods are inherently task-specific, demonstrate poor generalizability, and often require dedicated model for each dataset. To leverage the versatile capabilities of recently remarkable large language models (LLMs), several endeavors have explored generative approaches to entity extraction. Yet, these approaches often fall short of the effectiveness of previouly sequence labeling approaches. In this paper, we utilize the open-sourced LLM LLaMA2 as the backbone model, and design specific instructions to distinguish between different types of entities and datasets. By combining the LLM's understanding of instructions with sequence labeling techniques, we use mix of datasets to train a model capable of extracting various types of entities. Given that the backbone LLMs lacks specialized medical knowledge, we also integrate external entity knowledge bases and employ instruction tuning to compel the model to densely recognize carefully curated entities. Our model VANER, trained with a small partition of parameters, significantly outperforms previous LLMs-based models and, for the first time, as a model based on LLM, surpasses the majority of conventional state-of-the-art BioNER systems, achieving the highest F1 scores across three datasets.



\end{abstract}


\end{frontmatter}


\section{Introduction}

The importance of Biomedical Named Entity Recognition (BioNER) is underscored by its applications in various areas. For instance, mining knowledge of biomedical literature, supporting the clinical decision or advancing of biomedical research. It also served as downstream tasks such as information retrieval, relation extraction or biomedical question answering.

The traditional paradigm in BioNER employs sequence labeling, relying on pre-trained language models like BERT\citep{2019_bert} to encode the input sentence. Such methods are typically trained on specific datasets, often with suboptimal outcomes. 
Some endeavors have investigated multi-task learning\citep{2019_mtl} (MTL), leveraging shared representations across multiple datasets to enhance overall performance. While a shared representation layer is employed, the decoding layers (e.g. Multilayer Perceptron or Conditional Random Field) are often task-specific, leading to a monolithic representation-based BioNER models that may underperform when confronted with out-of-domain tasks \citep{2023_zero}. 
However, sharing decoding layers also presents issues, since BioNER datasets are curated according to different standards. Merely concatenating these datasets yields inconsistent annotations where the same entity is tagged in one dataset but omitted in another, compromising dataset quality and thereby model performance.

Recently, generative pre-trained transformer (GPT) models have achieved state-of-the-art (SOTA) performance across multiple natural language processing (NLP) tasks. 
These large language models excel at adhering to instructions through fine-tuning with task-specific directives. By incorporating distinct task descriptions into the input instruction, they effectively harness the power of LLMs for instruction tuning on multiple datasets, thereby offering a solution to the aforementioned challenges. This approach not only imparts the models with rich representations but also enables them to reconcile discrepancies among different datasets. Currently, research in the NLP community is heavily focused on decoder-based LLMs, while investigation into encoders has begun to lag behind. Addressing how to more effectively employ decoder-based LLMs for entity extraction is therefore an urgent priority.

However, performance in information extraction tasks, such as relation extraction and entity extraction, falls short of conventional state-of-the-art (SOTA) methods. These tasks typically involve sequence labeling, where studies\citep{2023_lsllama,2024_notright} have identified challenges related to the causal mask (CM) in the transformer-based attention mechanism. The CM, which restricts bidirectional information flow during training, can hinder token annotation influenced by surrounding tokens. Removing this mask during training and inference can significantly enhance model performance.

Based on this, we adopt the open-source LLaMA2 model \citep{2023_llama2} as the backbone of our BioNER system, removing the causal mask and adopting a sequence labeling approach for entity extraction. Here, the LLM serves as a high-performance feature extractor capable of understanding instructions.
However, BioNER requires domain-specific knowledge, which general LLMs like LLaMA is trained on general corpus, lacking deep understanding of the biomedical domain.
Incorporating multiple BioNER datasets during training can alleviate this problem by facilitating knowledge transfer across domains. Furthermore, we introduce an efficient method for injecting external medical knowledge using knowledge bases such as UMLS\citep{2004_umls}, thereby enhancing model understanding of bio-entities. 
This approach, along with multi-datasets instruction tuning, enriches the model's information processing capabilities.

\begin{figure*}[h]
    \centering
    \includegraphics[width=2.0\columnwidth]{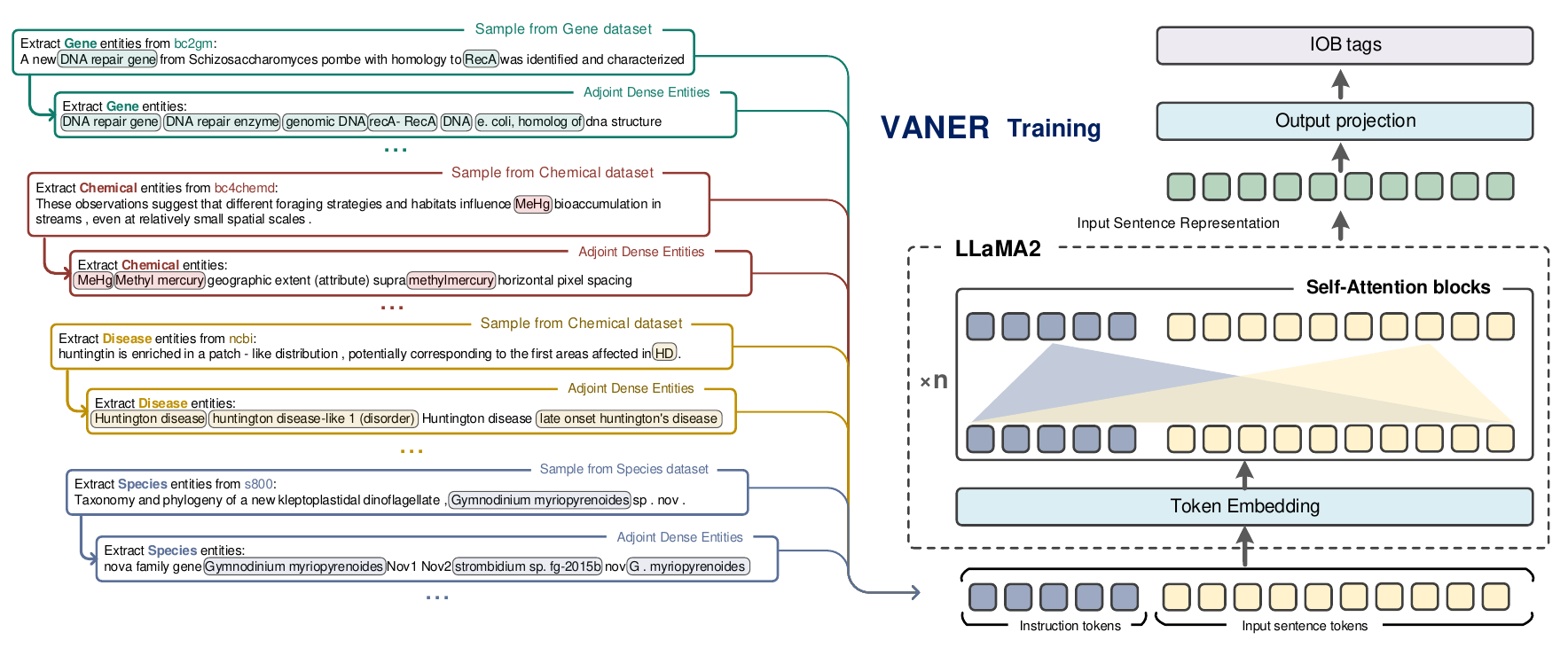}
    \caption{Training framework of VANER, accomplished through instruction tuning with a mixture of the four types of entities depicted on the left. Shaded areas in the sentences on the left indicate entities of corresponding types.}
    \label{fig_framework}
\end{figure*}


In summary, the contributions of this paper are as follows:
\begin{enumerate}
    \item We propose VANER, a unified BioNER solution that utilizes instruction tuning on LLMs and sequence labeling to handle diverse entities across multiple datasets. 
    \item We introduce a training method named Dense Bioentities Recognition (DBR) for injecting external knowledge, improving model convergence and performance.
    \item VANER achieves SOTA results on multiple datasets, and demonstrating robust domain adaptation capabilities. 
    \item Our method is resource-efficient, requiring only a single 4090 GPU for training and inference. We make the code, data, and pre-trained models publicly available\footnote{https://github.com/Eulring/VANER}.
\end{enumerate}


\section{Related Work}

\subsection{Biomedical Named Entity Recognition}

In the BioNER field, the majority of approaches adopt sequence labeling. Sequence labeling necessitates encoding the input sequence, with early models being LSTM-CRFs \citep{2017_habibi}, originally proposed by \citep{2015_BiLSTMCRF}, which utilize BiLSTM \citep{1997_LSTM} for encoding and subsequently combine it with CRF \citep{2001_crf} to predict the tag sequence that identifies the entities. Subsequently, the models for encoding input sequences transitioned to BERT \citep{2019_bert} and its biomedical domain-specific variants, such as BioBERT \citep{2020_biobert}, PubMedBERT \citep{2021_pubmedbert}, or BiolinkBERT \citep{2022_linkbert}.


The biomedical domain possesses numerous datasets, and jointly training on these datasets can enhance model performance. For example, \citet{2019_mtl} utilized multi-task learning to augment model encoding capabilities. AIONER \citep{2023_aioner} integrated labels from distinct datasets for unified extraction. KGQA \citep{2021_kgqa} extracted entities from various datasets via question answering, leveraging external knowledge as context. HUNER \citep{2020_huner} consolidated training on 34 datasets to boost model performance in particular categories. HunFlair\citep{2021_hunflair}, building upon HUNER, with the approach involving pretraining on vast amounts of biomedical text prior to fine-tuning on numerous BioNER datasets.

\subsection{Large Language Model for NER}


LLMs have achieved remarkable results across numerous tasks without supervised fine-tuning. However, for entity extraction, certain prompt-based methods, like GPT-NER\citep{2023_gptner} or PromptNER\citep{2023_promptner}, have shown poor performance, particularly on biomedical entity datasets. \citet{2023_inspire} aims to enhance the performance of LLMs by retrieving external medical knowledge. However, their effectiveness falls far short compared to supervised methods.


In supervised approaches, through instruction tuning\citep{2021_IT2,2024_T5,2022_IT1}, UIE\citep{2023_uie} and InstructUIE\citep{2023_iuie} handle various information extraction tasks using the same generative model. Following this, similar methods include UniversalNER\citep{2023_uniner}, which uses LLaMA2\citep{2023_llama2} as the backbone and distills knowledge from GPT-4, enabling it to extract entities from multiple datasets simultaneously. 
BioNER-LLaMA\citep{2024_bionerllama}, which draws inspiration from the output format of GPT-NER\citep{2023_gptner}, and is a generative, bio-medical domain-focused, fine-tuning based approach.

To better adapt LLMs to NER tasks, some researches\citep{2023_lsllama,2024_notright,2024_llm2vec} has found that removing the causal mask from the decoder during the training and inference phases of LLMs can be advantageous for sequence labeling tasks. VANER adopts this concept and further refines it with improvements tailored to the biomedical domain.

\section{VANER}

\subsection{Task Definition}

We define a set of biomedical NER datasets as \(D=\{D_1, D_2, ..., D_M\}\). Each dataset \(D_i\) corresponds to a name \(N_i\) and a category \(T_i\) respectively. 
In the dataset \(D_i\), it contains the input sentences \(X = \{x_1, x_2, ..., x_n\}\), as well as the corresponding entity tags \(Y = \{y_1, y_2, ..., y_n\}\).
Since we use the IOB tagging scheme, each label \(y_i\) belongs to \(\{I, O, B\}\). The objective of BioNER is to train one or more NER taggers that can extract entities of specified category \(T_i\) from the input sentence \(X\).

\subsection{NER as Instruction Tuning}
Instruction tuning \cite{2022_IT1,2021_IT2} is a method that enables a Large Language Model (LLM) to comprehend natural language instructions and generate tailored responses. Through instruction tuning, LLMs can produce outputs tailored to various requirements, thereby enhancing the model's multifunctionality. 

When designing instructions for the sentence \(X\) from the dataset \(D_i\), we provide the dataset name \(N_i\) as well as the desired entity type \(T_i\). 
Here, \(T_i\) guides the LLMs on what type of entities to output, while \(N_i\) offers information about the dataset. Different datasets may have variations in annotation rules and normalization, so informing the dataset name allows the LLM to adjust its output according to the dataset's style. The resulting instruction is denoted as \(\text{Instruct}(N_i, T_i)\), and when combined with the sentence \(X\), it forms the input \(I_X\):

\begin{align}
    I_{X} = \text{Instruct}(N_i, T_i) \oplus X
\end{align}

The specific form of the instruction can refer to Figure \ref{fig_framework}.

\subsection{LLM for Sequence Labeling}
In VANER, we utilize the LLM to understand instructions and simultaneously generate the input token embeddings \(H_X \in \mathds{R}^{n \times dim}\). Here, \(dim\) represents the dimension of the hidden state within the LLM, and \(n\) is the length of the input sentence \(X\).
\begin{align}
    H_{X} = \text{LLaMA2\_embedding}(I_{X})
\end{align}
This differs from traditional decoders that produce outputs in an autoregressive manner. Additionally, during the computation of self-attention, following \cite{2023_lsllama,2024_notright,2024_llm2vec} we remove the causal mask in the attention calculation. The feature representation $H_{X}$ is then projected into the IOB label space and the probability \(P \in \mathds{R}^{n\times 3}\) for each tag is calculated:
\begin{align}
    P = \text{softmax} \left( W H_{X} + b \right)
\end{align}
During training, the loss for input \(X\) is calculated using the cross-entropy loss:
\begin{align}
    Loss(X, Y) = CE(P, Y)
\end{align}

\begin{figure}[h]
    \centering
    \includegraphics[width=\linewidth]{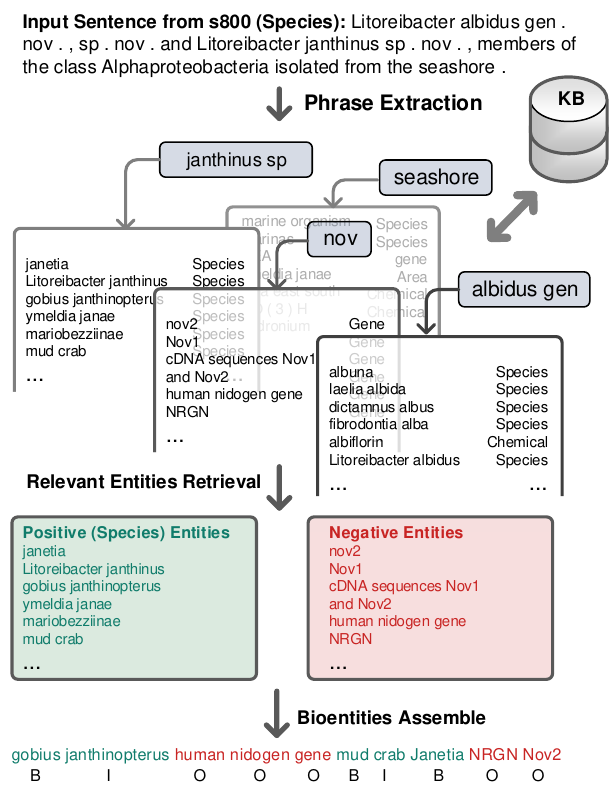}
    \caption{Production of training samples for dense entity recognition.} 
\label{fig:dbr}
\end{figure}

\subsection{Dense Bioentities Recognition (DBR)}

Biomedical entity recognition is a knowledge-driven task, and it can be beneficial for an LLM to retain objective entity knowledge. Previous methods\cite{2020_biobert,2021_pubmedbert,2022_linkbert} introduced biomedical knowledge inefficiently by pre-training on large-scale biomedical corpora or related biomedical tasks. In VANER, we enhance LLM's memory by selecting entities from external knowledge bases (e.g. UMLS\cite{2004_umls}) and densely concatenating them into a sentence with training instructions. This method enables the LLM to memorize entities directly.

At the same time, we believe that having the model memorize all entities from the knowledge base is inefficient in terms of training. We selectively choose high-quality entities from external knowledge bases for the LLM to remember, and consider entities that appear in the training sentence (possibly as negative entities) and entities similar to them as high-quality entities. Memorizing these entities is more conducive to enhancing the model's performance.
To facilitate this, we designed a process to retrieve relevant entities for each input sentence $X$ from dataset $D_i$.

\begin{itemize}
    \item \textbf{Step 1 Prepare the external knowledge base:} The external knowledge base consists of tuples comprising entity names and entity categories: 
    \begin{align}
    \mathcal{KB} = \left \{  <e_{\text{name}}^j, e_{\text{type}}^j> \big| 1\leq j \leq |\mathcal{KB}|    \right \}
    \end{align}

    \item \textbf{Step 2 Phrase Extraction:} Use an entity fragment mining tool AutoPhrase\citep{2018_autophrase} to extract potential entity fragments from input \(X\). Filter these fragments to obtain phrases for querying.
    
    \item \textbf{Step 3 Relevant Entity Retrieval:} For each filtered phrase, use it as a query to retrieval entities from knowledge base \(\mathcal{KB}\). We use the BioEntity representation model SapBERT\citep{2020_sapbert} to encode the query phrase and \(e_{\text{name}}\) in \(\mathcal{KB}\), and then calculate the cosine similarity. For each query, we retain the top $K$ relevant entities.

    \item \textbf{Step 4 Bio-Entities assemble:} 
    Retrieved entities encompass positive entities with \( e_{\text{type}} = T_i \), as well as negative entities. These are sampled and constituting a maximum of $t$ entities inclusive of both positive and negative samples, \( E = \{e_1, e_2, ..., e_t\} \). Concatenating these entities forms dense entities sentence \(X^* = \{x_1^*, x_2^*, ... , x_N^* \} \) for input.
    
\end{itemize}
    
By specifying the type \(T_i\) for the current dataset, we can obtain the corresponding instruction for \(X^*\). Since the entities are constructed from an external knowledge base, there is no need to specify the name of the dataset in the instruction:
\begin{align}
    I_{X^*} = \text{Instruct}(T_i) \oplus X^*
\end{align}

Based on the retrieved entity type and the expected output category \(T_i\) in the instruction, the training label \(Y^* = \{ y_1^*, y_2^*,... y_N^* \}\) for input \(X^{*}\) can be obtained as follows: 
\begin{equation}
    y^*_k =
    \left\{             
        \begin{array}{lr}             
        B, & \text{if } x_k^* \text{ is head token of } e \in E \text{ and } e_{\text{type}} = T_i\\
        I, & \text{if } x_k^* \text{ is inter token of } e \in E \text{ and } e_{\text{type}} = T_i\\
        O, & \text{otherwise}
        \end{array}
    \right.
\end{equation}

Similarly to the input \(X\) process, we obtain the instruction representation \(H_{P_e}\), the output label distribution \(d_e\), and the loss function:
\begin{align}
    H_{X^*} &= \text{LLaMA\_embedding}(I_{X^*}) \\
    P^* &= \text{softmax} \left( W H_{X^*} + b \right) \\
    {Loss}(X^*, Y^*) &= CE(P^*, Y^*)
\end{align}

\subsection{Multi-dataset Instruction Tuning}


By instruction tuning across multiple biomedical NER datasets and integrating Dense Bioentities Recognition, it comes to a unified loss $Loss_{\text{uni}}$, it is expressed as:
\begin{align}
    Loss_{\text{uni}} = \sum_{D_i \in D} \sum_{X \in D_i} Loss(X, Y) + Loss(X^*, Y^*) 
\end{align}
The resulting model, VANER, is capable of recognizing multiple types of bioentities with high performance.








\begin{table*}[h] 
  \centering
  \caption{Precision (P), Recall (R), and F1 scores(F1) on supervised BioNER datasets. $\dag$ represents the model using parameter-efficient fine-tuning. $*$ means the model is not finetuning based. The best F1 score in each dataset is \textbf{bolded}, and the second-best is \underline{underlined}.} \label{tab:main}
  \vspace*{0.4cm}
  \begin{tabular}{l l l P{7mm}P{8mm}P{8mm}P{8mm}P{9mm}P{8mm}P{9mm}P{7mm}P{8mm}}
    \toprule
    \multirow{2}{*}{Method} & \multirow{2}{*}{Backbone} & \multirow{2}{*}{Metric} & \multicolumn{2}{c}{Disease} & \multicolumn{2}{c}{Gene} & \multicolumn{2}{c}{Chemical} & \multicolumn{2}{c}{Species} & avg. \\
              &  &  &  ncbi  &bc5cdr  & bc2gm  &  jnlpba & bc4chem & bc5cdr & linnaeus & s800 & diff.  \\
    \midrule \midrule
     & \multicolumn{10}{c}{Traditional BioNER approaches} \\ \midrule
    BioBERT & BioBERT&F1& 88.12 & 83.83  & 81.94 & 76.23  & 89.55 & 91.98  & 88.17 & 71.26 & -3.03\\ 
    PubMedBERT &PubMedBERT&F1& 86.92 & 86.07  & 83.36 & 76.19  & 89.91 & 93.60  & 88.14 & 69.64 & -2.68 \\ 
    AIONER &BioBERT&F1& \underline{89.59} & \textbf{87.89} & - & - & - & 92.84 & \underline{90.63} & \textbf{79.67} & -0.16\\
    KGQA &BERT+CNN &F1& \textbf{89.69} & - & 83.14 & 79.24 & 91.82 & - & 90.61 & - & -0.33\\
    CompactBioBERT &BioBERT &F1& 88.67 & 85.38 & \textbf{86.71} & \underline{79.88} & 91.40 & 94.31 & 82.90 & 75.70 & -1.29\\
    BioLinkBERT & BioLinkBERT &F1& 88.76 & 86.39 & \underline{85.18} & \textbf{80.06} & - & \underline{94.04} & - & - & 0.68\\
    \midrule \midrule
     & \multicolumn{10}{c}{LLM-based approaches} \\ \midrule
    GPT-3.5 & $\text{GPT-3.5}^*$ &F1& 50.49 & 51.77 & 37.54 & 41.25 & - & 60.30  &  - & - & -37.93\\ 
    UnivresalNER &LLaMA2  &F1& 86.96 & - & 82.42 & 76.60 & 89.21 & - & - & - & -2.02\\
    InstructionUIE &T5 &F1& 86.21 & - & 80.69 & - & 87.62 & - & - & - & -3.39 \\
    BioNER-LLaMA2 &LLaMA2 &F1& 88.00 & - & 83.40 & - & 92.80 & - & - & - &  -0.16\\
    LS-unLLaMA &$\text{LLaMA2}^{\dag}$ &F1& 86.40 & 87.28 & 84.16 & 76.06 & 92.33 & 93.35 & 83.59 & 66.79 & -3.17\\  \cmidrule{2-12}
    \multirow{3}{*}{$\text{VANER}_{Single}$ (our)}
          &  &P& 88.73 & 86.57 & 83.48 & 70.74 & 93.20 & 93.61 & 86.64 & 72.04 & \\ 
          & $\text{LLaMA2}^{\dag}$ &R& 89.47 & 88.54 & 85.07 & 82.71 & 91.87 & 94.00 & 89.61 & 79.37 & \\ 
          &  &F1& 89.10 & 87.54 & 84.27 & 76.26 & \underline{92.53} & 93.80 & 88.10 & 75.53 & -1.02 \\ \cmidrule{2-12}
    \multirow{3}{*}{VANER (our)}
         &  &P& 87.13 & 86.70 & 83.18 & 71.99 & 93.64 & 93.98 & 93.61 & 73.75 & \\
         & $\text{LLaMA2}^{\dag}$ &R& 88.95 & 88.83 & 83.75 & 83.71 & 92.73 & 94.69 & 94.53 & 80.68 & \\
         &  &F1& 88.03 & \underline{87.76} & 83.47 & 77.41 & \textbf{93.18} & \textbf{94.34} & \textbf{94.06} & \underline{77.06} & - \\ 
    \bottomrule
  \end{tabular}
\end{table*}

\section{Experiments}

\subsection{Datasets}

In this paper, we combine and evaluate 8 datasets covering 4 entity categories, which include JNLPBA \citep{2004_jnlpba} and BC2GM \citep{2008_bc2gm} for gene types; Linnaeus \citep{2010_linnaeus} and S800 \citep{2013_s800} for species types; BC4CHEMD \citep{2015_bc4chemd} and BC5CDR-chem \citep{2016_bc5cdr} for chemical types; NCBI \citep{2014_ncbi} and BC5CDR-disease \citep{2016_bc5cdr} for disease types. Comprehensive statistics and detailed data regarding these datasets are presented in Table \ref{tab:dataset}.


To further validate the model's performance on unseen datasets, we assessed the adaptation capabilities of the trained VANER model on the CRAFT\citep{2017_craft} dataset. 
The CRAFT dataset encompasses three distinct entity classes: species, gene, and chemical. The number of instances for evaluation is 26,589.

\begin{table}[h] 
  \centering
  \caption{The statistics of the BioNER datasets used to train VANER.}
  \vspace*{0.4cm}
 \label{tab:dataset}
  \begin{tabular}{p{22mm}P{27mm}p{10mm}p{10mm}}
    \toprule
    Dataset Name & Dataset Splits & Types & Entities \\
    \midrule
    BC4CHEMD     & 30884 / 30841 / 26561 & Chemical & 65238\\
    s800         & 5743  / 831   / 1630 & Species & 3734\\
    linnaeus     & 12004 / 4086  / 7181 & Species & 2724\\
    JNLPBA       & 14731 / 3876  / 3873 & Gene & 30263\\
    BC2GM        & 12632 / 2531  / 5065 & Gene & 24453\\
    NCBI         & 5432  / 923   / 942  & Disease & 6861\\
    BC5CDR-disease         & 4559  / 4580  / 4796  & Disease & 12931\\
    BC5CDR-chem        & 4559  / 4580  / 4796  & Chemical & 15828\\
    \bottomrule
  \end{tabular}
\end{table}

\subsection{Baselines}
To evaluate the performance of the VANER in biomedical entity recognition tasks, we compared it with traditional methods and recent LLM-based approaches.

\subsubsection{Traditional BioNER baselines}

\paragraph{BioBERT\citep{2020_biobert}} is a pre-trained language model based on BERT\citep{2019_bert}, specifically designed for biomedical text mining tasks. It improves text comprehension and performance in biomedical domains through pre-training and fine-tuning on large-scale biomedical literature. 

\paragraph{PubMedBERT\citep{2021_pubmedbert}}
Similar to BioBERT, PubMedBERT is a BERT-based pre-trained language model. However, it primarily trains texts from the PubMed literature corpus.

\paragraph{AIONER\citep{2023_aioner}}
combines labels from multiple datasets to recognize multiple entities simultaneously. It integrates 11 BioNER datasets and uses PubMedBERT and BioBERT as the encoders for the model.

\paragraph{KGQA\citep{2021_kgqa}}
is trained on 18 datasets and treats  BioEntity extraction as a multi-answer question answering task, leveraging knowledge as context. It adopts BERT-CNN as its foundational architecture.

\paragraph{CompactBioBERT\citep{2023_combiobert}}
employs a blend of DistilBERT and TinyBERT distillation techniques derived from the BioBERT teacher model.

\paragraph{BioLinkBERT\citep{2022_linkbert}} is a BERT-based model pre-trained by leveraging links information between biomedical documents.

\subsubsection{LLM-based baselines}

We also compare our model against several LLM-based models described below.

\paragraph{GPT-3.5\citep{2023_gpt4}} conduct zero-shot inference by using OPENAI API. We directly borrow the results from \citep{2023_extensive}.

\paragraph{UniversalNER\citep{2023_uniner}}
is an LLM-based NER model distilled from ChatGPT, utilizing LLaMA2 as its backbone and undergoing full parameter fine-tuning. The model is trained on data from 43 datasets across 9 distinct domains.

\paragraph{InstructUIE\citep{2023_iuie}}
tackles information extraction tasks through instruction tuning, employing FlanT5-11B \citep{2024_T5} as its framework and undergoing full parameter fine-tuning. It utilizes 32 diverse information extraction datasets in a unified text-to-text format, accompanied by expert-authored instructions.

\paragraph{LS-unLLama\citep{2023_lsllama}}
is an LLaMA2-based sequence labeling approach. This method does not incorporate instructions into the input and is efficient fine-tuned on a single dataset. 

\paragraph{BioNER-LLaMA2\citep{2024_bionerllama}}
a generative model with LLaMA2 as its backbone, having full-parameter fine-tuning with a blend of three BioNER datasets.

\subsubsection{VANER variants}
As for the original version of VANER, we have set up some variants as controls for ablation and for different usage.

\paragraph{$\text{VANER}_{single}$}
For each dataset, $\text{VANER}_{single}$ is trained and tested independently using dataset-specific training data without combining data across multiple datasets.

\paragraph{VANER + random DBR}
During Dense Biomedical Recognition, entities are randomly selected from $\mathcal{KB}$ and integrated into the training process of VANER.

\paragraph{VANER w/o DBR}
VANER is trained without the utilization of Dynamic Biomedical Relation (DBR), focusing solely on the intrinsic learning from the dataset without additional entity integration.

\paragraph{$\text{VANER}_{adapt}$}
In instructions construction, only entity categories are included, without specifying the names of any entity datasets. The purpose of this setup is to allow the model to be more focused on its performance on unseen datasets rather a specific dataset.

\subsubsection{Off-the-shelf BioNER tools}
Performance on unseen datasets serves as a testament to the practicality of BioNER, leading us to select previously established SOTA off-the-shelf BioNER tools for comparison.

\paragraph{SciSpacy\citep{2019_scispacy}} is a specialized NLP library for processing biomedical text, using CNN to obtain token representations and employing POS taggers and dependency parsers.

\paragraph{HUNER\citep{2020_huner}} employs an LSTM-CRF architecture, trained on a consolidation of 34 datasets covering 5 entity types: cell line, chemical, disease, gene, and species. Annotators are trained separately for each of these 5 categories.

\paragraph{HunFlair\citep{2021_hunflair}} uses similar datasets to HUNER, but leverages character-level Flair embeddings and fastText embeddings, pre-trained on 3 million full biomedical texts and 25 million abstracts of biomedical articles. The model architecture employs a sequence labeling BiLSTM-CRF.

\subsection{Experiment Settings}

We optimized LLaMA2 via 4-bit LoRA\citep{2021_lora} fine-tuning, incorporating LoRA adapters into all linear layers with hyperparameters $r=16$ and $\sigma=32$. During training, the batch size was set to 8, and the input token length was limited to 128. The model underwent 20 epochs of training. The optimization procedure utilized the AdamW optimizer \citep{2018_adamw}, employing default momentum settings and a constant learning rate schedule fixed at $2 \times 10^{-4}$. VANER and its variants were executed on a single GTX-4090 GPU with 24GB of memory. Training VANER on a mixture of 8 datasets required 70 hours.

\begin{figure}[h]
\centering
\includegraphics[width=\linewidth]{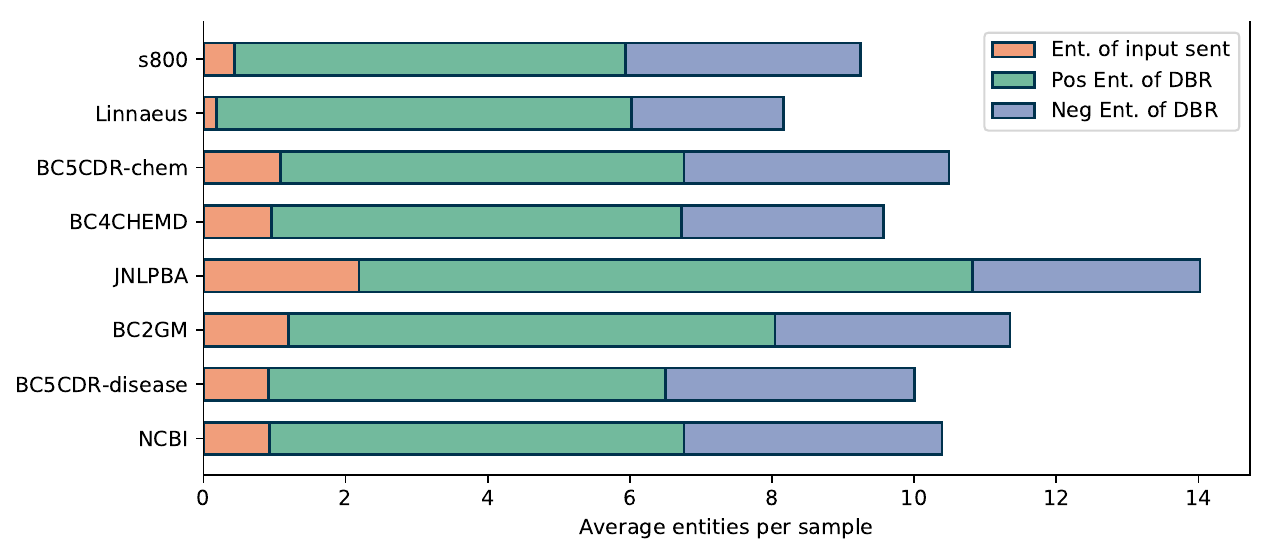}
\caption{Average entities count per sample on different datasets.}
\vspace*{0.4cm}
\label{fig:sta1}
\end{figure}

Entities of external knowledge base $\mathcal{KB}$ were mainly sourced from UMLS\citep{2004_umls} and part from the train split of datasets in Table\ref{tab:dataset}, with the $\mathcal{KB}$ housing 1 million entities. 
In the DBR assembling, a maximum of 10 positive entities are sampled for each sentence, and a maximum of 5 negative entities are sampled. 
Figure \ref{fig:sta1} illustrates the average number of entities contained in each input sentence across various datasets. For comparative purposes, the average number of positive entities and negative entities corresponding to the input's dense entities is also presented. It can be observed that the knowledge density associated with dense entities recognition significantly surpasses that of the original datasets.


\begin{table*}[ht] 
  \centering
  \caption{Ablation study of VANER.} 
  \vspace*{0.4cm}
  \label{tab:aba}
  \begin{tabular}{l l P{7mm}P{8mm}P{8mm}P{8mm}P{9mm}P{8mm}P{9mm}P{7mm}P{6mm}}
    \toprule
    \multirow{2}{*}{Method} & \multirow{2}{*}{Metric} & \multicolumn{2}{c}{Disease} & \multicolumn{2}{c}{Gene} & \multicolumn{2}{c}{Chemical} & \multicolumn{2}{c}{Species} & avg.  \\
                &  &  ncbi  &bc5cdr  & bc2gm  &  jnlpba & bc4chem & bc5cdr  &  linnaeus & s800 & diff. \\
    \midrule
  
     \multirow{3}{*}{$\text{VANER}_{Single}$}
        &P& 88.73 & 86.57 & 83.48 & 70.74 & 93.20 & 93.61 & 86.64 & 72.04 & \\ 
        &R& 89.47 & 88.54 & 85.07 & 82.71 & 91.87 & 94.00 & 89.61 & 79.37 & \\ 
        &F1& \textbf{89.10} & \textbf{87.54} & \textbf{84.27} & \textbf{76.26} & \textbf{92.53} & \textbf{93.80} & 88.10 & \textbf{75.53} & - \\  \cmidrule{2-11}
    \multirow{3}{*}{\ \ \ + random DBR}
        &P& 85.98 & 85.26 & 83.00 & 70.46 & 92.00 & 93.69 & 90.70 & 70.43 & \\ 
        &R& 87.59 & 87.58& 84.95 & 82.76 & 91.41 & 93.13 & 93.69 & 77.42 &  \\  
        &F1& 86.78 & 86.41 & 83.96 & 76.12 & 91.70 & 93.41 & \textbf{92.17} & 73.76 & -0.35\\ \cmidrule{2-11}
    \multirow{3}{*}{\ \ \  w/o DBR}
&P& 83.65 & 85.36 & 82.44 & 70.74 & 92.70 & 93.44 & 88.89 & 64.58 & \\  
        &R& 87.49 & 88.13 & 83.90 & 82.71 & 91.75 & 93.58 & 81.35 & 69.97 & \\  
        &F1& 85.52 & 86.72 & 83.16 & \textbf{76.26} & 92.22 & 93.51 & 84.95 & 67.17 & -2.20 \\
        
        \midrule

    \multirow{3}{*}{VANER}
    &P& 87.13 & 86.70 & 83.18 & 71.99 & 93.64 & 93.98 & 93.61 & 73.75 & \\  
    &R& 88.95 & 88.83 & 83.75 & 83.71 & 92.73 & 94.69 & 94.53 & 80.68 & \\  
    &F1& 88.03 & \textbf{87.76} & \textbf{83.47} & \textbf{77.41} & \textbf{93.18} & \textbf{94.34} & \textbf{94.06} & \textbf{77.06} & - \\  \cmidrule{2-11}
    \multirow{3}{*}{\ \ \ + random DBR}
        &P& 88.31 & 85.90 & 82.97 & 70.95 & 92.70 & 93.81 & 92.74 & 71.48 & \\ 
        &R& 89.05 & 88.38 & 83.08 & 82.77 & 91.13 & 94.41 & 94.11 & 76.89 &  \\ 
        &F1& 88.68 & 87.12 & 83.03 & 76.41 & 91.91 & 94.11 & 93.42 & 74.09 &  -0.81\\ \cmidrule{2-11}
    \multirow{3}{*}{\ \ \ w/o DBR}
    &P& 88.36 & 86.28 & 82.90 & 70.92 & 92.83 & 93.91 & 91.14 & 74.05 & \\ 
    &R& 90.20 & 88.35 & 83.92 & 83.24 & 91.04 & 93.49 & 94.46 & 78.59 & \\ 
        &F1& \textbf{89.27} & 87.30 & 83.41 & 76.59 & 91.93 & 93.70 & 92.77 & 76.25 & -0.51\\
    \bottomrule
  \end{tabular}
\end{table*}


\subsection{Main Results}
Table \ref{tab:main} presents the results of our models VANER and $\text{VANER}_{single}$ across 8 datasets, offering an extensive comparison with alternative methods. VANER achieves state-of-the-art performance on three datasets: BC4CHEMD, BC5CDR-chem, and Linnaeus. 
At the same time, we compared the average F1 scores of VANER and other models on available datasets. VANER outperformed all models except for BioLinkBERT.

Overall, apart from VANER, LLM-based approaches notably underperform compared to traditional BioNER methodologies. While LS-unLLaMA performs well on certain datasets, it exhibits poor performance on the two species datasets linnaeus and s800, possibly due to LLaMA2's lack of domain-specific biomedical knowledge. In contrast, our $\text{VANER}_{single}$ enhances training by incorporating DBR and appending instruction prompts to the input, resulting in significant improvement.

AIONER and KAGQ models achieve overall performance closely matching that of our VANER, with average F1 scores trailing by 0.16 and 0.33 respectively. Both models also employ training on a blend of multiple datasets, supporting the effectiveness of combining multiple datasets for training.

\begin{figure}[h]
\centering
\includegraphics[width=\linewidth]{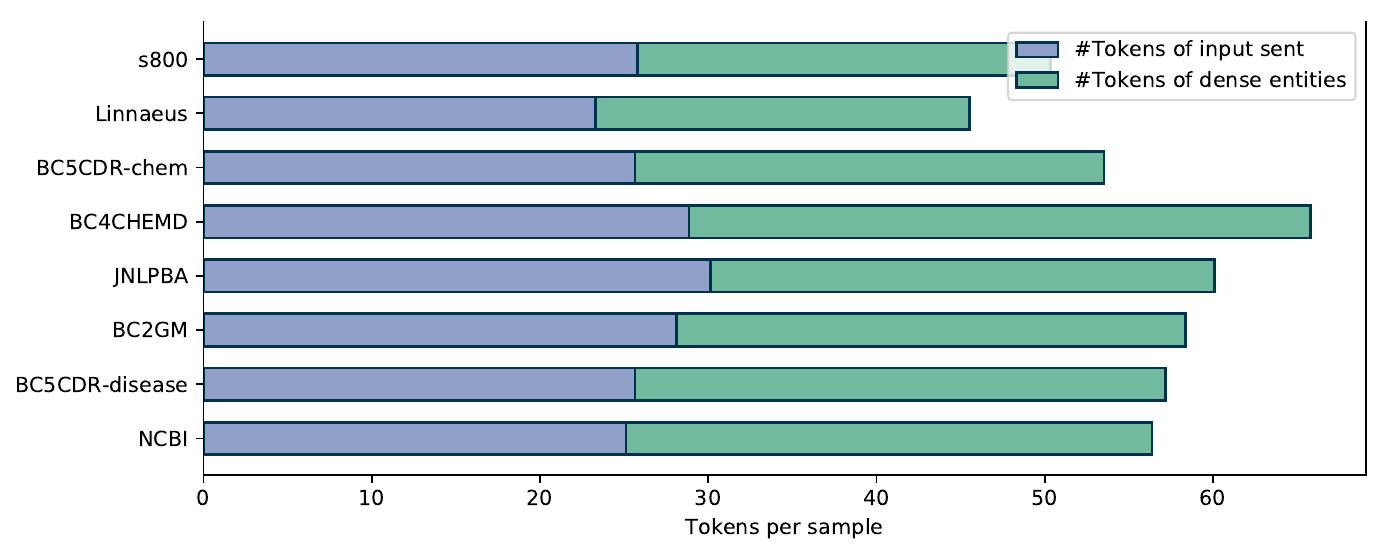}
\caption{Average token length of DBR and corresponding input sentence.}
\vspace*{0.4cm}
\label{fig:sta2}
\end{figure}
\begin{figure}[h]
  \centering
  \begin{minipage}{0.22\textwidth}
    \centering
    \includegraphics[width=\textwidth]{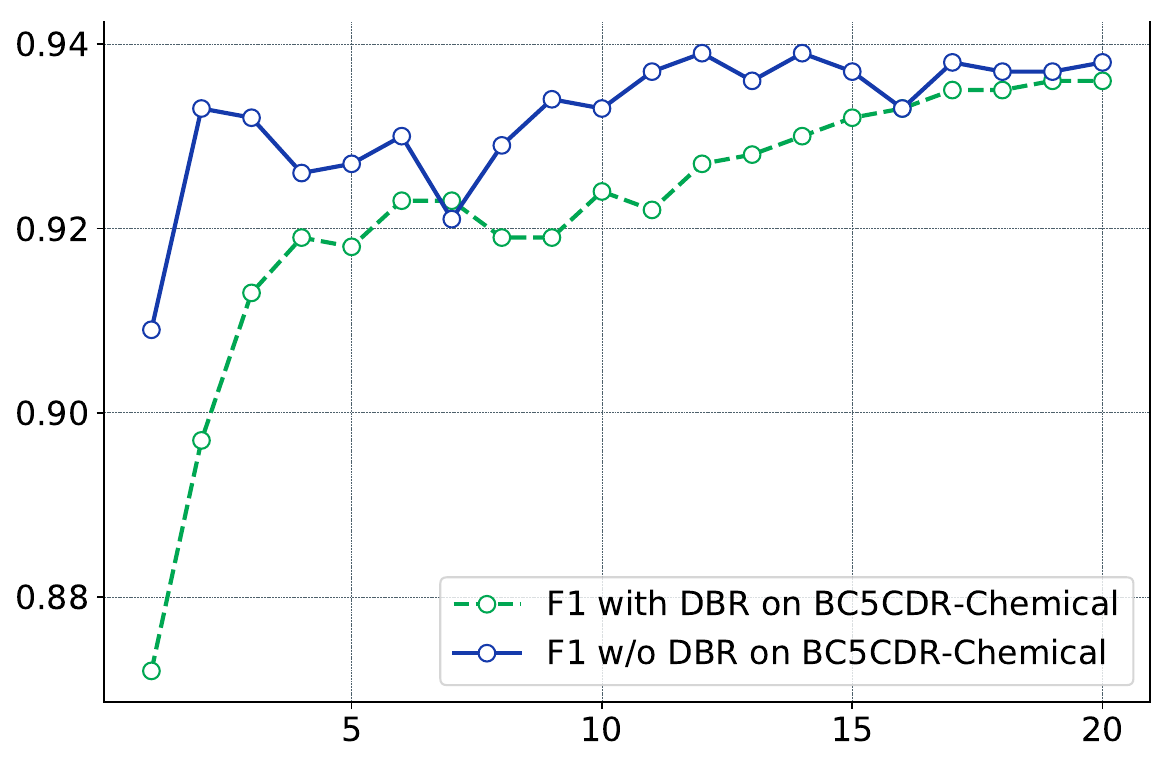}
    \label{fig:image1}
  \end{minipage}
  \begin{minipage}{0.22\textwidth}
    \centering
    \includegraphics[width=\textwidth]{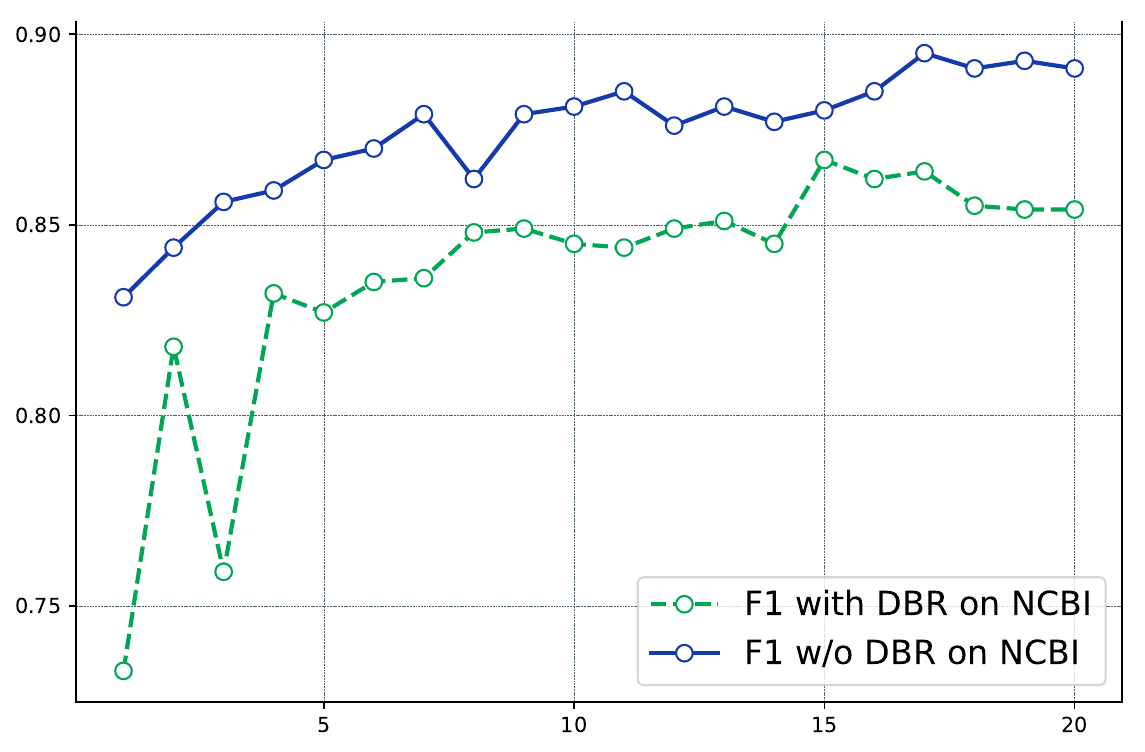}
    \label{fig:image2}
  \end{minipage}
  \caption{Varying the use of DBR in VANER training.}
  \vspace*{0.4cm}
  \label{fig:f1}
\end{figure}


\subsection{Ablation Study}

\subsubsection{Effect of Dense Bioentities Recognition}

Table \ref{tab:aba} presents findings from the ablation study concerning DBR. The entity knowledge integrated into both VANER and $\text{VANER}_{single}$'s DBRs is meticulously curated. In contrast, the random DBR method randomly samples from the $\mathcal{KB}$, controlling for the number of entity samples to match that of the DBR. Models are trained with identical architectures and hyperparameters to ensure a fair comparison. It is evident that random DBR falls short of DBR in performance, particularly on the NCBI dataset, where $\text{VANER}_{single}$ outperforms $\text{VANER}_{single}$+random DBR by 3 points. This indicates DBR's ability to inject more relevant and beneficial knowledge, thereby enhancing model performance. However, on the Linnaeus dataset, $\text{VANER}_{single}$ (F1=88.10) significantly underperforms $\text{VANER}_{single}$+random DBR (F1=92.17), suggesting that external entity knowledge proves more advantageous for this particular dataset.

Injecting corresponding entity knowledge into training samples on $\text{VANER}_{single}$ models yields notable improvements on NCBI, BC2GM, and JNLPBA, among others. Of particular note, on the NCBI dataset, F1 rises from 85.52 to 89.05. 
This demonstrates that without any additional data integration but solely by augmenting pertinent entity knowledge, $\text{VANER}_{single}$ can also have substantial progress, potentially reaching SOTA outcomes within its respective domain. 
Figure \ref{fig:f1} also charts the evolution of model performance during training for $\text{VANER}_{single}$ and the corresponding $w/o DBR$ method across two datasets, evidencing an accelerated convergence rate following DBR adoption.
Figure \ref{fig:sta2} also tallies the number of tokens in the DBR training instructions and the corresponding number of tokens in the input sentences. The difference between the two is not significant, indicating that $\text{VANER}_{single}$ achieves and even surpasses the performance of $w/o DBR$ with only a small amount of computational resources.

\subsubsection{Effect of Multi-dataset Instruction Tuning}
Table \ref{tab:aba} reveals performance enhancements following DBR implementation across models. Notably, the boost observed in $\text{VANER}_{single}$ surpasses that in VANER, indicating that both Multi-dataset Instruction Tuning and DBR inherently equip models with extrinsic knowledge beyond a single dataset. When no other dataset knowledge is available, the impact of DBR becomes more pronounced.

Additionally, a discernible trend is that the performance of VANER + random DBR dips below the average F1 score of VANER w/o DBR. This suggests that DBR imposes a higher standard on the quality of dense entities, such that incorporating low-quality ones under the Multi-dataset instruction tuning framework effectively introduces noise.

\begin{table}[h] \small
  \centering
\caption{The F1 scores on the CRAFT dataset.}
\vspace*{0.4cm}
  \begin{tabular}{l|P{11mm}P{11mm}P{11mm}}\toprule
        Method / Type     &  Species  &  Gene  &  Chemical      \\ \midrule 
        SciSpacy          & 54.21 & 47.76 & 35.73    \\
        HUNER             & 84.45 & 50.77 & 42.99   \\
        HunFlair          & 85.05 & \textbf{72.19} & \textbf{59.69}   \\ 
        VANER             & \underline{86.36} & 48.46 & 41.01   \\
        $\text{VANER}_{adapt}$   & \textbf{87.08}  & \underline{59.95} & \underline{48.12}   \\
 \bottomrule
  \end{tabular}
   \label{tab:adapt}
\end{table}

\subsection{Adaptation}


To validate the generalizability of the model, we conducted evaluations on the unseen CRAFT dataset. We selected VANER, VANER w/o DBR, as well as $\text{VANER}_{adapt}$ specifically designed for the general domain, and compared them against existing BioNER extraction tools, namely SciSpacy, HUNER, and HunFlair. The CRAFT dataset encompasses annotations for the entity types Species, Gene, and Chemical. Since the data set used for training VANER encompasses these three categories, it is suitable for extracting such entities. In the input instructions for all VANER models, no specific dataset name is included, only the required entity categories.


Table \ref{tab:adapt} presents the experimental outcomes, revealing that our model achieves superior performance on the species entity class. However, it falls short of HunFlair in the remaining two categories. Of particular note, our model displays less robust performance in recognizing gene entities. This disparity can be primarily attributed to the fact that HunFlair capitalizes on an extensive array of 13 gene datasets for its training regimen, while HUNER further consolidates learning from an additional 14 gene datasets. In contrast, VANER has been trained utilizing only two Gene datasets, despite our efforts to enhance dense bioentity recognition through the integration of external knowledge base information concerning entity profiles. This disparity in dataset size and quality underscores the critical role played by expanding the number of training resources, and concurrently, highlights the considerable scope for improvement inherent in our methodology.




\begin{figure}[h]
\centering
\includegraphics[width=\linewidth]{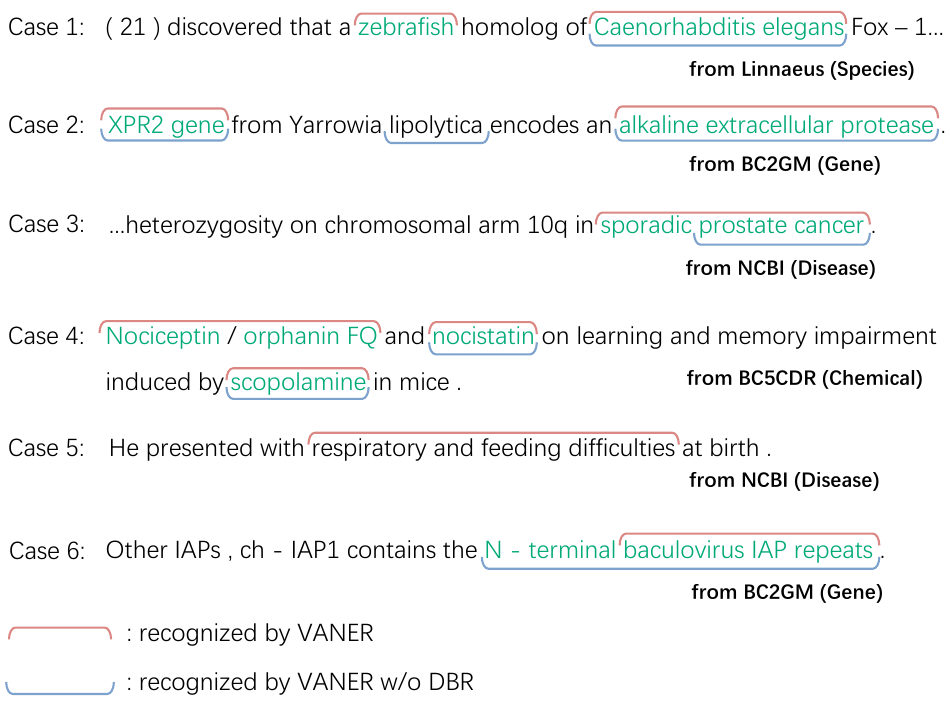}
\caption{Cases of VANER and VANER w/o DBR. The green tokens represent the golden entities.}
\vspace*{0.4cm}
\label{fig:cases}
\end{figure}
\subsection{Case Study}


In Figure \ref{fig:cases}, cases 1-3 demonstrate accurate predictions made by VANER. We have selected VANER without DBR for comparison. In the first case, VANER accurately identified `zebrafish' as a Species entity. In the second case, VANER without DBR incorrectly recognized `lipolytica' as a Gene entity. In fact, `lipolytica' is a Species entity, and without external domain knowledge, it is challenging to accurately identify this entity. In the third case, VANER correctly identified the boundaries of the entities.


In cases 4-6, several examples of incorrect predictions made by VANER are presented. In case 4, VANER treated `Nociceptin/orphanin FQ' as a single entity, whereas they are actually two distinct entities. In case 5, VANER misidentified a person's symptom as a disease. The cause of this may be the ambiguous delineation between the types of symptoms and diseases. In case 6, VANER incorrectly identified the boundaries of the entities. It can be observed that incorrect determination of entity boundaries and false recalls are relatively common errors in VANER. Additionally, as indicated in Table \ref{tab:main}, VANER has a higher recall rate on almost all datasets. This suggests that VANER rarely misses required entities, which also benefits from its extensive domain knowledge.

\section{Conclusion}


This paper introduces VANER, a novel versatile and adaptive BioNER model that integrates LLMs. VANER employs instruction tuning to train on multiple datasets. Unlike generative LLMs, VANER still utilizes sequence labeling to extract entities but relies on LLMs to understand instructions and generate vector representations of input tokens. We also designed a densen bio-entities recognition task to further enhance VANER's performance. Based on LLaMA2, VANER significantly outperforms previous LLM-based methods and approaches the effectiveness of state-of-the-art traditional BioNER methods through parameter-efficient fine-tuning across eight datasets. This bridges the gap between LLM-based methods and traditional approaches in NER tasks. By setting instructions, VANER can extract various categories of entities from a single dataset, demonstrating its versatility as a BioNER tool. Moreover, VANER shows excellent adaptation performance, accurately recognizing entities in some unseen datasets. The experimental results indicate that VANER is a viable and promising solution for BioNER.


VANER holds considerable room for improvement. By incorporating additional datasets and leveraging more powerful open-source LLMs, its performance can be further enhanced, presenting a viable avenue for future research. Moreover, while VANER is currently applied solely in the realm of entity extraction, we envision its integration with LLM capabilities enabling its application across a broader range of domains.

\bibliography{main}

\end{document}